\documentclass[conference]{IEEEtran}
\IEEEoverridecommandlockouts    

\usepackage{multirow}
\usepackage{amsmath}
\usepackage{amssymb}
 \usepackage[ruled,vlined]{algorithm2e}
\usepackage{graphicx}
\usepackage[hidelinks]{hyperref}
\usepackage{url}
\usepackage{cite}
\usepackage{booktabs}
 
\title{\vspace{0.15in}How Good Are Time-Series Foundation Models for Pedestrian Crowd Count Forecasting? A Cross-Dataset Comparative Study}

\author{
	\parbox{\textwidth}{%
		\centering
		Theivaprakasham Hari\textsuperscript{*}, Ziteng Li\textsuperscript{*}, Yanan Xin, Winnie Daamen, Serge Hoogendoorn%
	}
	\thanks{The authors are with the Department of Transport \& Planning, Faculty of Civil Engineering and Geosciences, Delft University of Technology, 2628 CN Delft, The Netherlands.}
    \thanks{\textsuperscript{*}These authors contributed equally to this work.}%
    \thanks{Corresponding author: Theivaprakasham Hari. Email: \texttt{T.Hari@tudelft.nl}.}%
}

\begin{document}
	
	\maketitle
	\thispagestyle{empty}
	\pagestyle{empty}
	

\begin{abstract}
Pedestrian-count forecasting supports pedestrian-oriented Intelligent Transportation Systems (ITS), including crowd monitoring, pedestrian-traffic staffing and routing, and proactive risk mitigation during surges.
Recent time-series foundation models (FMs) report strong zero-shot accuracy on heterogeneous forecasting benchmarks, but it remains unclear whether these gains transfer reliably to pedestrian sensing deployments. We benchmark seven univariate forecasting approaches spanning four paradigms: Seasonal Naive, gradient-boosted trees (LightGBM, CatBoost), deep learning models (N-HiTS, PatchTST), and two pretrained FMs (TimesFM, Chronos-2). Experiments cover two complementary regimes: (i)
a five-day special event dataset SAIL2025 at 3-minute resolution with limited in-domain history; and (ii) Melbourne pedestrian sensors as a multi-year hourly dataset (2010--2017) with strong seasonality. We compare the MAE and RMSE results per sensor across datasets and multiple forecast horizons. Results show three consistent findings. First, with limited historical data, Seasonal Naive remains a strong baseline for long-horizon forecasting on high-volume sensors, while trained models can degrade when the next day differs substantially from prior days. Second, boosted trees can be competitive on lower-volume sensors but exhibit higher sensitivity on high-volume sensors under event-driven shift. Third, FMs excel in the seasonal and data-rich regime under long-context configuration. The findings highlight the importance of choosing pedestrian forecasting models based on both the underlying data conditions and the forecasting horizon.
\end{abstract}

	\section{Introduction}
	\label{sec:introduction}

Pedestrian sensing networks are increasingly used in Intelligent Transportation Systems (ITS) to support operational planning, including crowd management, staffing, and routing \cite{duives2018monitoring, duives2020enhancing}. Forecasting in this domain is challenging because three factors often co-occur in practice: (i) limited local history during short deployments such as public events, 
(ii) day-to-day distribution shift driven by schedules, weather, and operational interventions \cite{mai2025learning}, and (iii) high short-term volatility, where pedestrian flows can change within minutes due to transit arrivals, bottlenecks, or local incidents, making near-term trajectories harder to extrapolate than in many vehicle-count settings~\cite{daamen2007flow}. Traditional time-series forecasting in ITS has relied on supervised machine learning (ML) and deep learning (DL) models trained on local historical data, with common baselines including Seasonal Naive \cite{hyndman2018forecasting}, direct multi-horizon gradient-boosted tree models (LightGBM and CatBoost) \cite{ke2017lightgbm,prokhorenkova2018catboost}, and specialized deep architectures for long-horizon forecasting (PatchTST and N-HiTS) \cite{nietime,challu2023nhits}. 

With the development of foundation models (FMs) and large language models (LLMs), time series foundation models have emerged as a potential alternative. Models such as Google's TimesFM and Amazon's Chronos-2 \cite{das2024decoder,ansari2025chronos}, pretrained on massive amounts of cross-domain sequence data, aim to learn general temporal representations, thereby supporting zero-shot or few-shot forecasts. In pedestrian sensing, this capability is most relevant for pop-up or newly instrumented locations, temporary event infrastructure, and rapidly reconfigured pedestrian management plans, where forecasts are needed before sufficient site-specific history accumulates for stable supervised training. 

While time-series foundation models theoretically offer advantages, systematic evidence remains limited on their actual gains over classic ML/DL models for pedestrian count forecasting. Their robustness under distribution shifts and their computational and latency implications are also important for deployment, especially under long-context inference. Since crowd monitoring systems typically run on modest on-premises hardware and must refresh forecasts for tens of sensors within each control cycle, latency, memory, and retraining cost matter alongside accuracy, yet benchmarks in this domain report error metrics almost exclusively. Current literature lacks a unified, reproducible, and engineering-usable benchmark covering diverse data conditions, often limiting conclusions to single data distributions or specific experimental setups.

To address this gap, we constructed and evaluated comparative experiments across data conditions using a unified data processing workflow and evaluation metrics, covering two datasets that instantiate two deployment regimes:
(i) a special event-driven, limited-history pedestrian deployment (SAIL2025); and
(ii) a long-running, highly seasonal urban sensing deployment (Melbourne). Through systematic comparisons with classical ML/DL baselines, we provide empirical evidence on when time-series FMs are useful for pedestrian-count forecasting and when simpler methods remain preferable. Beyond accuracy rankings, we emphasize deployment-relevant factors, including distribution shift, available history, and short- versus long-context inference. Our main contributions are as follows:

\begin{itemize}
\item We evaluate two state-of-the-art time-series foundation models, TimesFM and Chronos-2, against ML (direct multi-horizon LightGBM and CatBoost) and DL (PatchTST and N-HiTS) baselines.
\item We use the SAIL2025 dataset (5 days, 3-minute resolution) with an expanding walk-forward protocol to assess the zero-shot capabilities of FMs under limited-history event conditions.
\item We assess FM context-window capacity on the multi-year Melbourne dataset by using the maximum supported input lengths, 8,192 tokens for Chronos-2 and 16,000 time steps for TimesFM, and by examining how extended context affects long-horizon forecasting up to 720 hours.
\item We quantify what each model costs to deploy on identical CPU-only hardware, covering training time, inference latency, memory footprint, and model size.

\end{itemize}

The remainder of this paper is organized as follows.
Section~\ref{sec:relatedwork} reviews related work.
Section~\ref{sec:methodology} defines datasets, task formulation, models, and experimental protocols.
Section~\ref{sec:results} reports results and discussion.
Section~\ref{sec:conclusion} concludes with implications, limitations, and future directions.
    
\section{Related Work}
\label{sec:relatedwork}

This section reviews prior work on pedestrian count forecasting, multi-horizon deep forecasting models for time series, and time-series foundation models, highlighting gaps that motivate our evaluation.
    
\subsection{Pedestrian Count Forecasting}
\label{subsec:pedestrainflowforecasting}

Pedestrian-count forecasting has been studied using regression models, tree ensembles, and sequence-based architectures, all of which exploit the strong periodicities and location-specific dynamics observed in urban pedestrian flows \cite{wang2017predicting,sohn2020laying,hari2026evaluation}.
However, a recurring limitation of the existing literature is that conclusions are typically drawn from a single dataset or a narrow set of models.
This makes it difficult to assess whether reported gains generalize across data regimes, such as event-driven data limitation with only a few days of observations versus multi-year deployments with stable seasonal structure.

\subsection{Multi-Horizon Deep Forecasting Models}
\label{subsec:advancedtimeseiresforecasting}
Pedestrian management requires forecasts across multiple operational horizons, from minute-scale interventions to hour-scale staffing and longer-term planning.
Modern deep learning architectures address these horizons by representing long temporal contexts and periodic structure more effectively than classical methods.
Temporal Fusion Transformers incorporate multi-horizon attention and static covariates \cite{lim2021temporal}.
Informer reduces the computational complexity of long-sequence modeling through sparse attention \cite{zhou2021informer}.
Autoformer introduces seasonal-trend decomposition with autocorrelation-based attention to capture long-range dependencies \cite{wu2021autoformer}.
PatchTST improves long-horizon accuracy by segmenting the input series into patches and applying channel-independent Transformer encoding \cite{nietime}.
Several alternative architectures without attention mechanisms also remain competitive. For instance, TiDE uses a dense multilayer perceptron (MLP) encoder-decoder with low inference overhead \cite{das2023long}.
These methods typically rely on sufficient in-domain history and stable periodic patterns, so their multi-horizon forecast performance can degrade during special events where pedestrian flows shift rapidly and only a short local observation window is available.
In summary, while these architectures are promising when stable periodicity and adequate training data exist, their benefits under short, shift-heavy pedestrian event windows remain uncertain.
This gap is particularly relevant because pedestrian event deployments, which are among the most operationally critical use cases for ITS, provide exactly the conditions where these models are least tested.
A controlled benchmark that evaluates models across both data-rich and limited-history regimes is therefore needed.

\subsection{Time-Series Foundation Models}
\label{subsec:timeseiresfoundationalmodels}
Time-series foundation models aim to learn transferable temporal representations through large-scale pretraining on heterogeneous corpora and to support zero-shot or few-shot forecasting with minimal task-specific adaptation.
LLM-inspired approaches treat forecasting as sequence completion or reprogram frozen language model backbones for temporal forecasting \cite{gruver2023large,jin2024time}.
Lag-Llama introduces a univariate probabilistic forecasting foundation model pretrained on a large corpus of time series from diverse domains \cite{rasul2023lag}.
TimesFM is a pretrained decoder-only Transformer designed for general-purpose forecasting with support for long context windows \cite{das2024decoder}.
Chronos-2 formulates forecasting as token prediction after discretizing real-valued series into a fixed vocabulary and reports strong zero-shot performance across diverse benchmark datasets \cite{ansari2025chronos}.
Although these models report competitive results on standard benchmarks such as Monash \cite{godahewa2021monash} and M4~\cite{makridakis2020m4}, their evaluation on pedestrian sensor data remains limited.
In particular, pedestrian forecasting during special events requires reliable forecasting under severely limited in-domain history and strong day-to-day distributional shifts, conditions that are not represented in existing FM benchmark suites.
FMs therefore offer an attractive training-free deployment path for pedestrian sensing, especially for temporary installations and newly instrumented event corridors where no site-specific history exists.
However, their reliability under these conditions, along with the practical value of long-context inference when stable periodicity is available, remains insufficiently documented.
Establishing this evidence is the central objective of our study.

\section{Methodology and Experiments}
\label{sec:methodology}

This section defines the forecasting task, describes the two datasets, and specifies the evaluated model families and their configurations.

\subsection{Problem Formulation}
\label{subsec:problem_formulation}
For each sensor, we model its readings as a univariate time series $\{y_t\}_{t=1}^{T}$ of pedestrian counts sampled at a fixed cadence $\Delta$.
Sensors are evaluated independently. In the event setting, the short deployment window and occasional missing or low-activity periods limit the amount of informative in-domain history. Given a context window of length $L$ (in time steps), the task is to forecast the next $H$ values:
\begin{equation}
\hat{\mathbf{y}}_{t+1:t+H} = f\!\left(\mathbf{y}_{t-L+1:t}\right),
\end{equation}
where $f(\cdot)$ denotes the forecasting model. We evaluate multiple forecast horizons $H$. For foundation models, we additionally vary $L$ to quantify the sensitivity of forecast accuracy to the amount of context available at inference time.

\subsection{Datasets}
\label{subsec:datasets}
We benchmark univariate pedestrian count forecasting on two datasets that reflect complementary deployment regimes.
SAIL2025 represents an event-driven setting with limited historical data and pronounced day-to-day shift.
Melbourne represents a data-rich, highly seasonal setting suitable for long-horizon evaluation. We treat each sensor channel as an independent univariate series and report results per sensor. This choice matches common deployment practice, where each sensor must remain functional as a standalone unit, and avoids requiring spatial metadata. Modeling cross-sensor dependence is a complementary extension beyond the scope of this study.

\subsubsection{Melbourne Pedestrian Dataset}
The Melbourne Pedestrian Counting System provides hourly pedestrian counts from automated sensors distributed across Melbourne's central business district~\cite{melbourne_pedcounts_com} (Table~\ref{tab:dataset_overview}, Fig.~\ref{fig:dataset_overview}).
We select 16 sensors that have continuous hourly observations from 2010 to 2017 with no missing values, yielding 70{,}128 timestamps per sensor.
The data exhibit strong daily and weekly seasonality as well as substantial scale differences across sensor locations, with mean hourly counts ranging from 147 (sensor T8) to 1{,}508 (sensor T4).
These characteristics make the Melbourne dataset a suitable benchmark for comparing model performance at long forecast horizons up to 720 hours.

\subsubsection{SAIL2025 Event Dataset}
\begin{figure}[ht]
    \centering
    \includegraphics[width=\columnwidth]{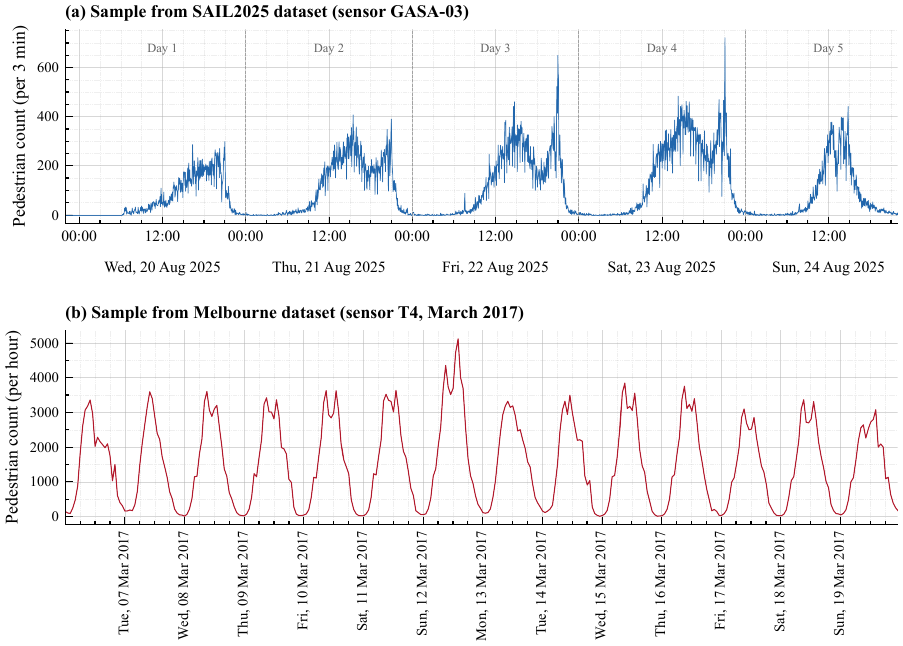}
    \caption{Overview of the two benchmark datasets. (a)~Sample from the SAIL2025 event dataset (sensor GASA-03, 3-minute resolution) showing five consecutive days. (b)~Sample from the Melbourne pedestrian dataset (sensor T4, hourly resolution) showing a 14-day window in March 2017, illustrating the strong intra-daily periodicity.}
    \label{fig:dataset_overview}
\end{figure}

We use 11 sensors from Amsterdam's Crowd Monitoring System\footnote{\url{https://maps.amsterdam.nl/lvma/}}, including sensors deployed on major pedestrian corridors and ferry terminals.
The five-day event period is recorded at 3-minute resolution, producing 480 samples per day and 2{,}400 timestamps per sensor.
Notably, the opening and closing days show systematically different temporal profiles than the intermediate days (Fig.~\ref{fig:dataset_overview}), intensifying the day-to-day distributional shift observed in the walk-forward evaluation setting.

\subsection{Forecasting Models}
\label{subsec:forecastingmodels}
We benchmark seven univariate forecasting models grouped into four categories (Table~\ref{tab:model_summary}).
All models produce point forecasts.
For Chronos-2, which generates probabilistic outputs, we use the median forecast (P50) to ensure comparability with the other models.

\subsubsection{Seasonal Naive}
Seasonal Naive repeats the most recent seasonal cycle, defined as $\hat{y}_{t+h}=y_{t+h-s}$, where $s$ is the seasonal period.
For SAIL2025, we set $s=480$ (one day at 3-minute resolution), and for Melbourne, we set $s=168$ (one week at hourly resolution).

\subsubsection{Gradient-Boosted Trees}
We evaluate LightGBM \cite{ke2017lightgbm} and CatBoost \cite{prokhorenkova2018catboost}. Both models are trained using a direct multi-horizon strategy that fits one independent regressor per forecast step $h\in\{1,\ldots,H\}$ to avoid recursive error accumulation.
For each forecast origin $t$, the input feature vector consists of all lagged counts within the input window, $\mathbf{x}_t = [y_t, y_{t-1}, \ldots, y_{t-L+1}]$, with $L=120$ timestamps for SAIL2025 and $L=168$ hours for Melbourne.
We include no exogenous covariates or calendar features, isolating the performance attributable to univariate history alone.
To maintain a controlled comparison across regimes, both models use a fixed configuration (100 estimators, maximum depth of 6, learning rate of 0.1) applied identically to both datasets.
This choice prioritizes interpretability of regime-specific effects over dataset-specific tuning. As a consequence, absolute errors for the tree baselines may be conservative relative to a fully optimized per-dataset configuration.

\subsubsection{Deep Learning Baselines}
We evaluate N-HiTS \cite{challu2023nhits} and PatchTST \cite{nietime} using standard untuned architectures to maintain consistency with the fixed-configuration design of the tree-based models.
For N-HiTS, we use three stacks with 512-unit hidden layers.
For PatchTST, we use three encoder layers, four attention heads, a patch length of 16, and a stride of 8.
Both models are trained independently per sensor with early stopping (patience of 5 epochs) and are capped at 500 gradient updates using the Adam optimizer with a learning rate of $10^{-3}$.

\subsubsection{Foundation Models}
TimesFM \cite{das2024decoder} and Chronos-2 \cite{ansari2025chronos} are applied strictly in zero-shot inference mode without any fine-tuning or parameter updates.
Both models operate directly on the context lengths specified in our experimental setups. Inference for both foundation models was run on an HP laptop equipped with a 13th-generation Intel Core i7-1355U CPU (x86-64, 12 logical cores) and 16 GB of RAM. No GPU acceleration was used.

\begin{table}[t]
\centering
\caption{Dataset overview and selected representative sensors.}
\label{tab:dataset_overview}

\small
\renewcommand{\arraystretch}{1.03}  
\setlength{\tabcolsep}{3pt}
\setlength{\aboverulesep}{0.2ex}
\setlength{\belowrulesep}{0.2ex}
\setlength{\cmidrulesep}{0.35ex}

\begin{tabular}{@{}lcc@{}}
\toprule
\textbf{Property} & \textbf{SAIL\,2025} & \textbf{Melbourne} \\
\midrule
Frequency   & 3 min & 1 hour \\
Coverage    & 5 days (Aug 2025) & 2010--2017 \\
Sensors     & 11 & 16 \\
Timestamps  & 2\,400 & 70\,128 \\
\midrule
\multicolumn{3}{@{}l}{\textit{Representative sensors (mean/max count, tier):}} \\
\;\;High   & GASA-03\_285\;(103/718)   & T4\;(1508/8052) \\
\;\;Medium & GASA-04\_135\;(64/291)    & T9\;(477/4272) \\
\;\;Low    & GASA-01-A2\_135\;(28/134) & T8\;(147/3009) \\
\bottomrule
\end{tabular}
\end{table}

\begin{table}[t]
\centering
\caption{Overview of benchmarked forecasting models.}
\label{tab:model_summary}

\small
\renewcommand{\arraystretch}{1.03}  
\setlength{\tabcolsep}{3pt}
\setlength{\aboverulesep}{0.2ex}
\setlength{\belowrulesep}{0.2ex}
\setlength{\cmidrulesep}{0.35ex}

\begin{tabular}{@{}llp{3.6cm}@{}}
\toprule
\textbf{Category} & \textbf{Model} & \textbf{Description} \\
\midrule
Baseline & Seasonal Naive & Repeats the most recent seasonal cycle~\cite{hyndman2018forecasting} \\
\midrule
\multirow{2}{*}{ML} & LightGBM & Gradient boosting with histogram-based splitting~\cite{ke2017lightgbm} \\
                   & CatBoost & Gradient boosting with ordered boosting~\cite{prokhorenkova2018catboost} \\
\midrule
\multirow{2}{*}{DL} & N-HiTS   & Hierarchical interpolation neural network~\cite{challu2023nhits} \\
                   & PatchTST & Patch-based Transformer~\cite{nietime} \\
\midrule
\multirow{2}{*}{FM} & TimesFM  & Decoder-only pretrained Transformer~\cite{das2024decoder} \\
                   & Chronos-2 & Tokenized T5-family language model~\cite{ansari2025chronos} \\
\bottomrule
\end{tabular}
\end{table}


\subsection{Experimental Setup 1: Event and limited-history regime (SAIL2025)}
\label{subsec:experimentsetupSAIL}
We evaluate day-by-day deployment readiness using an expanding walk-forward protocol that simulates real-time event operations.
In experiment $E_k$ ($k\in\{1,2,3,4\}$), all ML and DL models are trained on observations from the first $k$ event days and evaluated on day $k{+}1$.
For example, E1 trains on Day~1 and tests on Day~2, while E4 trains on Days~1--4 and tests on Day~5.
This protocol yields four successive train-test evaluations, progressively increasing the available training data from one to four days.
For the foundation models, we consider input context lengths of $L\in\{120,480\}$ timestamps, corresponding to 6 hours and 24 hours respectively, to quantify sensitivity to available history.
Forecast horizons are set to $H\in\{40,120,240,480\}$ timestamps, corresponding to 2, 6, 12, and 24 hours ahead.
The trained baselines (LightGBM, CatBoost, N-HiTS, PatchTST) are fitted using only the training days available in each walk-forward step.
Foundation models (TimesFM, Chronos-2) are applied in zero-shot mode and do not update any parameters.

\subsection{Experimental Setup 2: Seasonal and data-rich regime (Melbourne)}
\label{subsec:experimentsetupMelbourne}

We use a fixed chronological split with non-overlapping years.
The training period spans 2010-01-01 to 2015-12-31, the validation period covers 2016-01-01 to 2016-12-31, and the test period covers 2017-01-01 to 2017-12-31.
All trainable models are fit exclusively on the training period.
Early stopping and checkpoint selection for the deep learning baselines use the validation year, and all final results are reported on the held-out test year.

We evaluate forecast horizons of $H\in\{24,168,720\}$ hours, corresponding to 1 day, 1 week, and 1 month ahead.
For the foundation models, we compare a short context window ($L=168$ hours) with each model's maximum supported context on this dataset.
Specifically, we evaluate TimesFM with $L=16{,}000$ hours and Chronos-2 with $L=8{,}192$ hours to quantify the effect of long-context inference on forecasting performance.

\subsection{Evaluation Metrics}
\label{subsec:evaluationmetrics}

We report Mean Absolute Error (MAE) and Root Mean Squared Error (RMSE), defined as follows:
\begin{equation}
\text{MAE} = \frac{1}{N}\sum_{i=1}^{N} |y_i - \hat{y}_i|,\quad
\text{RMSE} = \sqrt{\frac{1}{N}\sum_{i=1}^{N} (y_i - \hat{y}_i)^2}.
\label{eq:metrics}
\end{equation}
Because pedestrian sensors can differ substantially in both scale and dynamics, we report metrics per sensor rather than aggregating across sensors into a single global score.

\subsection{Computational Cost Protocol}
\label{subsec:costprotocol}

We profile each model on the hardware described in Section~\ref{subsec:forecastingmodels}, recording training wall-clock time where applicable, the peak resident set size above the post-load baseline reached during fitting and during serving, the serialized model size, and the latency of a single $H$-step forecast.
Latency is taken as the median over 20 consecutive origins after two warm-up calls and is measured through each library's ordinary serving interface, so that it reflects the delay an operator actually experiences rather than an isolated forward pass; every model receives the same $L$ observations of context, and each is profiled in a separate process so that memory retained by previously loaded models cannot inflate the peak.
We profile one representative configuration per data regime on a high-volume sensor of each dataset, namely SAIL2025 at $L=120$ and $H=120$ (walk-forward step E4) and Melbourne at $L=168$ and $H=168$.

\section{Results and Discussion}
\label{sec:results}

This section reports qualitative and quantitative results for each dataset in turn, followed by a cross-regime comparison and discussion of model selection and deployment implications.

\subsection{SAIL2025 Event Dataset}
\label{subsec:result-sail2025}

\subsubsection{Qualitative Analysis}
\begin{figure}[ht]
    \centering
    \includegraphics[width=\columnwidth]{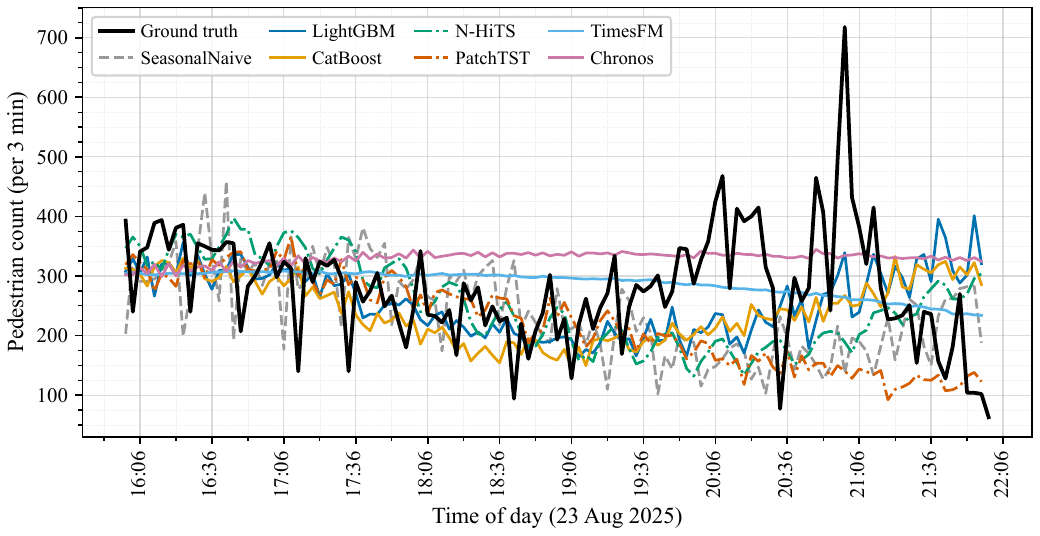}
    \caption{SAIL2025 dataset qualitative example on sensor GASA-03\_285: 6-hour-ahead forecast ($H=120$ at 3-minute resolution) during 16:00--22:00 on the test day of the E3 walk-forward step (trained on Days 1--3 and tested on Day 4; 23 August 2025).}
    \label{fig:sail_qualitative}
\end{figure}

Fig.~\ref{fig:sail_qualitative} compares 6-hour-ahead forecasts on the SAIL2025 high-volume sensor during the evening of Day~4 (23 August 2025).
The ground truth shows a post-peak decline followed by a pronounced resurgence between 20:00 and 21:20 driven by fireworks programming.
Seasonal Naive reproduces the previous day's shape and captures the coarse declining trend but misses the resurgence; tree-based models produce smoother trajectories, while deep learning models track the decline more closely.
Both FMs over-smooth the forecasts, with Chronos-2 failing to represent the downward trajectory entirely.
None of the seven models captures the late-evening crowd surge, highlighting that schedule-driven events with no training-set precedent cannot be predicted from the univariate series alone.

\subsubsection{Quantitative Results}
Table~\ref{tab:sail_walkforward} reports MAE across four walk-forward experiments as training data expands from one day (E1) to four days (E4).
On the high-volume sensor, Seasonal Naive achieves the lowest 24-hour MAE of 89.3 in E4, outperforming all learned models by at least 20.2 points, while LightGBM leads at the 2-hour horizon in three of four steps.
On medium- and low-volume sensors, gradient-boosted trees dominate: CatBoost achieves the best E4 MAE at all four horizons on the medium-volume sensor (14.9 to 40.6) and the best 24-hour MAE on the low-volume sensor (16.6).
Deep learning models provide mixed gains; PatchTST records competitive short-horizon accuracy in selected steps but does not generalize consistently.
FMs are most competitive in E1, where Chronos-2 achieves the best 2-hour MAE across all three sensors (35.9, 21.5, and 11.2) without any prior event data, confirming zero-shot cold-start value.
This advantage diminishes as local training data accumulates.

\begin{table*}[ht]
\centering
\caption{SAIL2025 dataset: Walk-forward progression of MAE across all forecast horizons. At the 3-minute sampling resolution, 2\,h, 6\,h, 12\,h, and 24\,h correspond to 40, 120, 240, and 480 timestamps, respectively. The table reports the 6\,h input-context setting, corresponding to $L=120$ timestamps. Training data grows from one day (E1) to four days (E4). For E1, trained baselines are reported only for the 2\,h and 6\,h horizons due to limited training data availability; zero-shot FM values are also shown at the 12\,h and 24\,h horizons where available. Best value per column within each horizon group is shown in bold. Second best is underlined.}
\label{tab:sail_walkforward}

\small
\renewcommand{\arraystretch}{1.03}   
\setlength{\tabcolsep}{3.0pt}        
\setlength{\aboverulesep}{0.2ex}
\setlength{\belowrulesep}{0.2ex}
\setlength{\cmidrulesep}{0.35ex}

\begin{tabular}{@{}cl rrrr rrrr rrrr@{}}
\toprule
 & & \multicolumn{4}{c}{\shortstack{\textbf{GASA-03\_285} \\ \textbf{(High volume)}}} &
     \multicolumn{4}{c}{\shortstack{\textbf{GASA-04\_135} \\ \textbf{(Medium volume)}}} &
     \multicolumn{4}{c}{\shortstack{\textbf{GASA-01-A2\_135} \\ \textbf{(Low volume)}}} \\
\cmidrule(lr){3-6} \cmidrule(lr){7-10} \cmidrule(lr){11-14}
\textbf{Horizon} & \textbf{Model} & E1 & E2 & E3 & E4 & E1 & E2 & E3 & E4 & E1 & E2 & E3 & E4 \\
\midrule

\multirow{7}{*}{2\,h} & Seasonal Naive  & 50.0 & 31.5 & 42.5 & 85.2 & \underline{21.8} & 20.9 & 35.1 & 46.8 & 14.2 & 13.5 & 16.7 & 18.5 \\
                      & LightGBM       & 63.9 & \textbf{27.1} & \textbf{37.0} & \textbf{35.6} & 26.9 & \textbf{17.5} & 31.9 & \underline{15.3} & 12.7 & \underline{12.3} & \underline{15.2} & \underline{12.3} \\
                      & CatBoost       & 72.1 & \underline{28.3} & \underline{39.5} & 41.1 & 26.1 & \underline{17.7} & 32.7 & \textbf{14.9} & 15.6 & \textbf{12.1} & \underline{15.2} & \textbf{11.6} \\
                      & N-HiTS         & 43.9 & 46.3 & \underline{39.5} & \underline{38.3} & 27.4 & 20.6 & \underline{27.9} & 15.8 & 80.2 & 19.6 & \textbf{15.1} & 14.9 \\
                      & PatchTST       & 87.2 & 59.6 & 52.0 & 40.6 & 29.0 & 20.7 & \textbf{27.8} & 17.3 & 20.0 & 16.2 & \textbf{15.1} & 12.4 \\
                      & TimesFM        & \underline{40.6} & 51.9 & 55.9 & 51.5 & 22.5 & 24.9 & 33.0 & 20.2 & \underline{11.5} & 14.1 & 16.1 & 13.3 \\
                      & Chronos-2      & \textbf{35.9} & 51.3 & 56.7 & 49.7 & \textbf{21.5} & 22.0 & 29.3 & 17.6 & \textbf{11.2} & 13.8 & 15.6 & 13.8 \\
\midrule

\multirow{7}{*}{6\,h} & Seasonal Naive  & \textbf{54.9} & \textbf{30.6} & \textbf{43.1} & 74.6 & \textbf{24.6} & \underline{22.8} & 39.2 & 50.3 & \underline{15.3} & 14.5 & 17.8 & 18.4 \\
                      & LightGBM       & 73.3 & \underline{30.8} & \underline{43.4} & \underline{69.3} & \underline{35.2} & \textbf{20.7} & \textbf{38.0} & \underline{23.7} & \textbf{14.6} & \underline{14.1} & \underline{17.2} & \underline{14.0} \\
                      & CatBoost       & 78.6 & \underline{30.8} & 45.7 & \textbf{68.1} & 36.2 & \textbf{20.7} & \underline{38.6} & \textbf{23.4} & 18.3 & \textbf{13.9} & \textbf{16.6} & \textbf{13.4} \\
                      & N-HiTS         & 84.7 & 64.8 & 59.5 & 71.1 & 50.8 & 47.4 & 50.2 & 31.5 & 35.6 & 35.4 & 21.3 & 28.7 \\
                      & PatchTST       & 94.3 & 83.4 & 71.1 & 80.8 & 42.8 & 32.7 & 52.6 & 39.5 & 31.6 & 29.3 & 21.9 & 18.8 \\
                      & TimesFM        & 73.1 & 98.8 & 105.2 & 102.3 & 40.5 & 47.4 & 67.2 & 40.6 & 15.9 & 21.3 & 27.6 & 25.7 \\
                      & Chronos-2      & \underline{62.1} & 94.3 & 99.3 & 115.6 & 40.2 & 41.6 & 60.1 & 36.0 & 15.5 & 20.3 & 26.0 & 26.0 \\
\midrule

\multirow{7}{*}{12\,h} & Seasonal Naive & -- & \textbf{29.9} & \textbf{43.4} & 65.4 & -- & 25.1 & \textbf{43.9} & 53.0 & -- & 16.0 & 19.6 & 19.4 \\
                       & LightGBM      & -- & 31.6 & \underline{46.9} & \textbf{62.2} & -- & \underline{23.2} & \underline{45.6} & \underline{35.3} & -- & \underline{14.9} & \underline{18.7} & \underline{16.4} \\
                       & CatBoost      & -- & \underline{31.1} & 48.1 & \underline{63.9} & -- & \textbf{23.1} & 45.7 & \textbf{32.7} & -- & \textbf{14.7} & \textbf{18.5} & \textbf{15.9} \\
                       & N-HiTS        & -- & 57.7 & 106.8 & 73.3 & -- & 37.4 & 63.0 & 44.0 & -- & 36.9 & 32.0 & 51.2 \\
                       & PatchTST      & -- & 108.1 & 104.1 & 73.6 & -- & 45.7 & 73.6 & 48.0 & -- & 36.3 & 29.1 & 23.2 \\
                       & TimesFM       & \underline{96.9} & 142.2 & 162.0 & 138.3 & \textbf{50.7} & 69.1 & 94.7 & 65.1 & \underline{20.4} & 30.2 & 38.9 & 35.0 \\
                       & Chronos-2     & \textbf{81.5} & 138.8 & 160.1 & 167.3 & \underline{54.8} & 67.3 & 88.5 & 59.1 & \textbf{20.0} & 29.0 & 37.1 & 34.1 \\
\midrule

\multirow{7}{*}{24\,h} & Seasonal Naive & -- & \underline{33.6} & \textbf{42.1} & \textbf{89.3} & -- & 19.9 & \underline{33.2} & 44.4 & -- & \underline{12.9} & \textbf{15.9} & 17.6 \\
                       & LightGBM      & -- & 34.1 & \underline{62.3} & 123.3 & -- & \textbf{18.7} & 34.0 & 44.8 & -- & 13.1 & 18.1 & \underline{17.2} \\
                       & CatBoost      & -- & \textbf{33.5} & 63.8 & 118.6 & -- & \underline{19.5} & \textbf{33.0} & \textbf{40.6} & -- & \textbf{12.5} & 18.7 & \textbf{16.6} \\
                       & N-HiTS        & -- & 83.5 & 100.2 & \underline{109.5} & -- & 31.6 & 36.0 & \underline{42.0} & -- & 13.8 & \underline{16.5} & 23.2 \\
                       & PatchTST      & -- & 59.8 & 79.2 & 112.8 & -- & 25.6 & 38.5 & 42.9 & -- & 20.8 & 22.7 & 18.2 \\
                       & TimesFM       & \underline{100.8} & 128.3 & 152.6 & 122.5 & \textbf{51.8} & 62.4 & 85.3 & 61.4 & \textbf{21.6} & 27.9 & 33.3 & 29.0 \\
                       & Chronos-2     & \textbf{88.9} & 143.9 & 175.3 & 174.8 & \underline{56.1} & 77.1 & 98.8 & 62.9 & \underline{22.0} & 29.2 & 34.5 & 31.7 \\
\bottomrule
\end{tabular}
\end{table*}

\subsection{Melbourne Pedestrian Dataset}
\label{subsec:result-melbourne}

\subsubsection{Qualitative Analysis}
\begin{figure}[ht]
    \centering
    \includegraphics[width=\columnwidth]{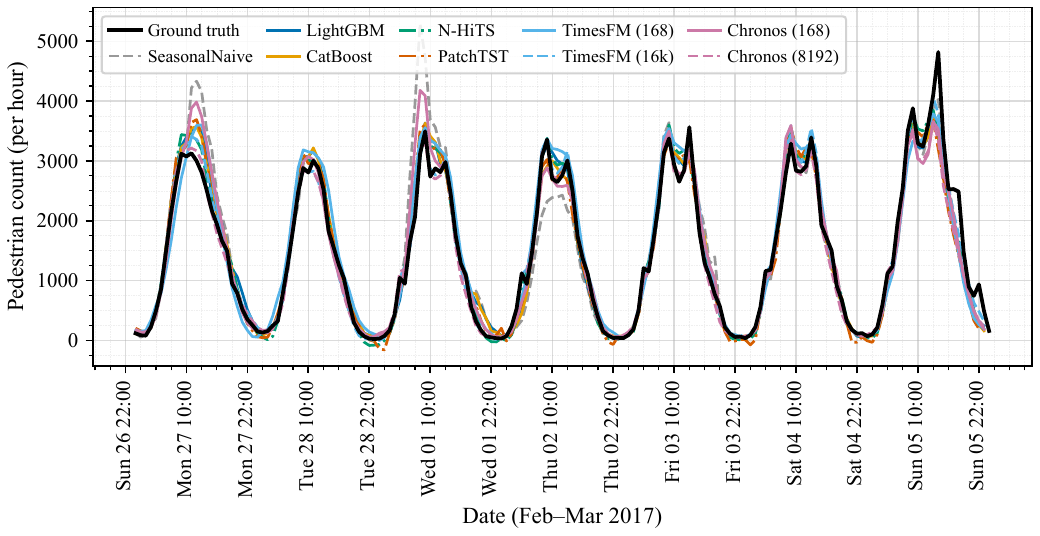}
    \caption{Qualitative one-week-ahead forecast comparison ($H=168$ at 1-hour resolution) on the Melbourne dataset (high-volume sensor T4) for the period 27~February--5~March~2017. The naive baseline, gradient-boosted trees, and deep-learning models use an input context length of $L=168$. Foundation models are evaluated with both short ($L=168$) and maximum supported long context windows (TimesFM: $L=16{,}000$; Chronos-2: $L=8{,}192$).}
    \label{fig:melb_qualitative}
\end{figure}

Fig.~\ref{fig:melb_qualitative} shows a one-week-ahead forecast on high-volume sensor T4 (27~February--5~March~2017).
The ground truth exhibits strong daily periodicity with weekday peaks substantially exceeding weekend activity.
Supervised models capture the weekday--weekend distinction but tend to under-predict the largest peaks.
Short-context FMs ($L=168$) produce reduced-amplitude forecasts, whereas extending context to $L=16{,}000$ (TimesFM) and $L=8{,}192$ (Chronos-2) visibly recovers peak magnitudes and daily rhythm, illustrating that longer context enables FMs to exploit the stable periodicity of multi-year pedestrian data.

\subsubsection{Quantitative Results}

Table~\ref{tab:melbourne_per_sensor} reports MAE and RMSE for forecast horizons of 24, 168, and 720 hours on three representative sensors.
Long-context inference produces consistent accuracy gains for both FMs.
TimesFM with $L=16{,}000$ hours reduces MAE by approximately 16--49\% relative to $L=168$ across the reported sensors and horizons, achieving the best or second-best MAE at the 168- and 720-hour horizons.
Chronos-2 with $L=8{,}192$ shows smaller but consistent improvements (approximately 12 to 13\% on T4 and T9 at 24 hours) and achieves the best MAE on the medium- and low-volume sensors at the daily and weekly horizons.
Context length is thus a primary factor governing FM performance on this dataset: the seasonal regime rewards models that can exploit long periodic structure.

\begin{table*}[ht]
\centering
\caption{Melbourne dataset: MAE and RMSE by sensor and forecast horizon. Non-FM models use an input window of 168\,h (1 week). Foundation models are evaluated with short context ($L$=168\,h) and long context (TimesFM: $L$=16{,}000\,h; Chronos-2: $L$=8{,}192\,h). The best value per column is shown in bold; the second-best value is underlined.}
\label{tab:melbourne_per_sensor}

\small
\renewcommand{\arraystretch}{1.03}   
\setlength{\tabcolsep}{2.6pt}        
\setlength{\aboverulesep}{0.2ex}
\setlength{\belowrulesep}{0.2ex}
\setlength{\cmidrulesep}{0.35ex}

\begin{tabular}{@{}l rr rr rr rr rr rr rr rr rr@{}}
\toprule
 & \multicolumn{6}{c}{\textbf{Sensor T4 (High Volume)}} &
   \multicolumn{6}{c}{\textbf{Sensor T9 (Medium Volume)}} &
   \multicolumn{6}{c}{\textbf{Sensor T8 (Low Volume)}} \\
\cmidrule(lr){2-7} \cmidrule(lr){8-13} \cmidrule(lr){14-19}
 & \multicolumn{2}{c}{\textbf{24\,h}} & \multicolumn{2}{c}{\textbf{168\,h}} & \multicolumn{2}{c}{\textbf{720\,h}}
 & \multicolumn{2}{c}{\textbf{24\,h}} & \multicolumn{2}{c}{\textbf{168\,h}} & \multicolumn{2}{c}{\textbf{720\,h}}
 & \multicolumn{2}{c}{\textbf{24\,h}} & \multicolumn{2}{c}{\textbf{168\,h}} & \multicolumn{2}{c}{\textbf{720\,h}} \\
\cmidrule(lr){2-3} \cmidrule(lr){4-5} \cmidrule(lr){6-7}
\cmidrule(lr){8-9} \cmidrule(lr){10-11} \cmidrule(lr){12-13}
\cmidrule(lr){14-15} \cmidrule(lr){16-17} \cmidrule(lr){18-19}
\textbf{Model}
& \scriptsize MAE & \scriptsize RMSE & \scriptsize MAE & \scriptsize RMSE & \scriptsize MAE & \scriptsize RMSE
& \scriptsize MAE & \scriptsize RMSE & \scriptsize MAE & \scriptsize RMSE & \scriptsize MAE & \scriptsize RMSE
& \scriptsize MAE & \scriptsize RMSE & \scriptsize MAE & \scriptsize RMSE & \scriptsize MAE & \scriptsize RMSE \\
\midrule
Seasonal Naive
& 201.4 & 360.6 & 203.0 & 362.9 & 237.8 & 406.3
& 113.6 & 343.3 & 110.5 & 337.6 & 126.3 & 369.4
& 40.7 & 87.8 & 40.6 & 88.1 & 44.1 & 94.8 \\
LightGBM
& 148.9 & \underline{252.8} & 169.7 & \underline{283.8} & 195.0 & \underline{316.0}
& 110.7 & 245.7 & 142.5 & 308.1 & 193.0 & 402.0
& 30.3 & \underline{61.9} & 34.5 & 69.0 & 39.8 & 77.0 \\
CatBoost
& 160.7 & 268.2 & 178.1 & 294.0 & 203.5 & 324.7
& 117.1 & 263.4 & 138.7 & 319.2 & 174.8 & 390.5
& 32.0 & 63.6 & 35.5 & 70.3 & 39.9 & 77.6 \\
N-HiTS
& 150.4 & 258.4 & 186.7 & 306.0 & 217.9 & 346.4
& 84.2 & 228.2 & 115.1 & 288.9 & 132.1 & 316.3
& 31.0 & 64.5 & 36.6 & 73.0 & 41.1 & 80.9 \\
PatchTST
& 194.9 & 296.1 & 214.2 & 328.1 & 248.1 & 367.5
& 119.3 & 284.3 & 132.7 & 323.8 & 147.8 & 356.0
& 34.7 & 68.9 & 37.1 & 74.0 & 41.1 & 81.4 \\
TimesFM ($L$=168)
& 197.3 & 310.5 & 202.4 & 329.0 & 230.4 & 364.8
& 141.2 & 292.7 & 130.2 & 296.3 & 145.4 & 333.6
& 36.0  & 68.9  & 38.0  & 74.6  & 42.4  & 82.2 \\
TimesFM ($L$=16{,}000)
& \textbf{143.4} & \textbf{243.2} & \textbf{154.5} & \textbf{259.9} & \textbf{178.8} & \textbf{297.2}
& \underline{71.4}  & \underline{200.8} & \underline{79.3}  & \textbf{232.1} & \textbf{97.4} & \textbf{265.5}
& \underline{28.7}  & 62.1  & \underline{31.1}  & \textbf{67.0}  & \textbf{35.5} & \textbf{73.6} \\
Chronos-2 ($L$=168)
& 167.6 & 293.2 & 189.3 & 323.3 & 224.6 & 371.8
& 78.0  & 215.6 & 90.0  & 259.5 & 113.3 & 306.4
& 29.5  & 62.8  & 34.2  & 71.5  & 40.1  & 81.4 \\
Chronos-2 ($L$=8{,}192)
& \underline{146.2} & 260.3 & \underline{166.7} & 287.4 & \underline{191.3} & 319.0
& \textbf{68.7} & \textbf{193.0} & \textbf{79.1} & \underline{235.4} & \underline{99.7}  & \underline{283.4}
& \textbf{27.0} & \textbf{59.9}  & \textbf{30.8} & \underline{67.9}  & \underline{35.7}  & \underline{76.1} \\
\bottomrule
\end{tabular}
\end{table*}

\subsection{Computational Cost and Deployment Latency}
\label{subsec:result-efficiency}

Table~\ref{tab:efficiency} reports the deployment cost of the benchmarked models. The Seasonal Naive model covers one seasonal cycle in a few kilobytes with inference in microseconds, whereas the trained baselines need 21 to 781\,s to fit and 100 to 1{,}392\,ms to produce a single forecast. The zero-shot FMs eliminate training cost entirely, although this advantage is accompanied by substantially larger storage and memory footprints. Chronos-2 requires 115\,ms per forecast on the SAIL dataset and 117\,ms on the Melbourne dataset, while occupying 875\,MB and 881\,MB memory space during inference, respectively, with a 478\,MB checkpoint. TimesFM, by contrast, requires 910\,ms per forecast on the SAIL dataset and 913\,ms on the Melbourne dataset, together with 961\,MB and 973\,MB of inference memory, respectively, and a 925\,MB checkpoint, the largest storage footprint among all evaluated models.
Since Seasonal Naive also achieves the lowest 24-hour MAE on the high-volume SAIL sensor in E4 (Table~\ref{tab:sail_walkforward}), such additional computational expense provides no benefit in terms of accuracy within the event regime.

LightGBM and CatBoost each accommodate a single regressor per forecast step, resulting in both training and inference processes that scale linearly with the horizon length $H$. Consequently, these models, despite being conceptually among the simplest, are also among the slowest trained models to serve, requiring between 299\,ms and 1{,}392\,ms, and have the largest storage requirements among the trained baselines, reaching up to 19.5\,MB on the SAIL dataset and 46.7\,MB on the Melbourne dataset.

In the case of Melbourne, CatBoost exemplifies an extreme scenario, necessitating 781 seconds for training and occupying 1.28\,GB during serving. This imposes limitations on the number of sensors that can be concurrently managed by a machine with 16\,GB of RAM. For the zero-shot FMs, the main deployment advantage is that they do not require repeated training, even though their memory usage is not necessarily lower.

The number of parameters is also an unreliable indicator of model efficiency. For instance, PatchTST contains approximately ten times fewer parameters than N-HiTS; however, it exhibits a training speed about six times slower. This disparity is attributable to the fact that attention mechanisms over patches are more computationally demanding per gradient step than hierarchical interpolation. Accordingly, N-HiTS emerges as the most economical model in terms of training within the seasonal regime. Simultaneously, it also constitutes the most robust supervised model at the 24-hour forecast horizon for sensor T9, as detailed in Table~\ref{tab:melbourne_per_sensor}.

Even so, these magnitudes remain compatible with real-time operation at the network sizes studied here. On the SAIL dataset, TimesFM, the slowest model at 910\,ms per forecast, would refresh all 11 sensors sequentially in approximately 10\,s within a 3-minute cycle. On the Melbourne dataset, CatBoost remains the slowest model and would refresh all 16 sensors in approximately 22\,s within an hourly cycle. For models requiring local fitting, the binding constraint is retraining, which costs roughly 675\,s for the four trained baselines at a single sensor and horizon and extrapolates to about two hours per walk-forward step across the full SAIL network, whereas the Seasonal Naive and the zero-shot FMs are exempt from it entirely.

\begin{table*}[ht]
\centering
\caption{Computational cost is measured under the protocol of Section~\ref{subsec:costprotocol}, including memory, latency, and model size. Zero-shot FMs incur no in-domain training cost.}

\label{tab:efficiency}

\small
\renewcommand{\arraystretch}{1.03}
\setlength{\tabcolsep}{3.0pt}
\setlength{\aboverulesep}{0.2ex}
\setlength{\belowrulesep}{0.2ex}
\setlength{\cmidrulesep}{0.35ex}

\begin{tabular}{@{}l rrrrr rrrrr@{}}
\toprule
 & \multicolumn{5}{c}{\shortstack{\textbf{SAIL2025} (GASA-03\_285) \\ $L$=120, $H$=120, 1{,}920 training samples}} &
   \multicolumn{5}{c}{\shortstack{\textbf{Melbourne} (T4) \\ $L$=168, $H$=168, 61{,}368 training samples}} \\
\cmidrule(lr){2-6} \cmidrule(lr){7-11}
\textbf{Model}
& \scriptsize\shortstack{Train\\(s)} & \scriptsize\shortstack{Train mem\\(MB)}
& \scriptsize\shortstack{Latency\\(ms)} & \scriptsize\shortstack{Infer. mem\\(MB)} & \scriptsize\shortstack{Size\\(MB)}
& \scriptsize\shortstack{Train\\(s)} & \scriptsize\shortstack{Train mem\\(MB)}
& \scriptsize\shortstack{Latency\\(ms)} & \scriptsize\shortstack{Infer. mem\\(MB)} & \scriptsize\shortstack{Size\\(MB)} \\
\midrule
Seasonal Naive
& $<$0.01 & $<$1 & $<$0.01 & $<$1 & 0.004
& $<$0.01 & $<$1 & 0.02 & $<$1 & 0.001 \\
LightGBM
& 21.2 & 333 & 299 & 308 & 19.5
& 384.6 & 860 & 910 & 333 & 46.7 \\
CatBoost
& 143.8 & 431 & 368 & 396 & 14.4
& 781.1 & 1{,}629 & 1{,}392 & 1{,}284 & 20.7 \\
N-HiTS
& 73.3 & 544 & 100 & 513 & 10.6
& 122.0 & 639 & 156 & 606 & 11.2 \\
PatchTST
& 436.6 & 775 & 178 & 613 & 1.0
& 686.6 & 957 & 163 & 843 & 1.4 \\
\midrule
TimesFM
& -- & -- & 910 & 961 & 925
& -- & -- & 913 & 973 & 925 \\
Chronos-2
& -- & -- & 115 & 875 & 478
& -- & -- & 117 & 881 & 478 \\
\bottomrule
\end{tabular}
\end{table*}

\subsection{Cross-Regime Comparison and Discussion}
\label{subsec:crossregime}
Comparing the two datasets reveals a fundamental regime dependence in model rankings.
In the event-driven regime, rankings are unstable across walk-forward steps, sensors, and horizons: Seasonal Naive leads at long horizons on the high-volume sensor, gradient-boosted trees dominate medium- and low-volume sensors, and FMs contribute primarily at cold start before local data accumulates.
In the seasonal regime, this pattern reverses: long-context FMs achieve the best or second-best MAE on all three Melbourne sensors at weekly and monthly horizons, while supervised models remain competitive at the daily scale.
No single model family is universally optimal; the deployment context, specifically the amount of available history and the degree of distributional shift, determines which approach is most effective.

Seasonal Naive's competitiveness on high-volume SAIL sensors reflects the fact that it makes no attempt to learn: copying the previous day's trajectory is a strong heuristic when consecutive event days share roughly similar crowd profiles, whereas learned models must estimate parameters from one to four days of observations, a sample size insufficient to capture the variance of high-volume event dynamics.
On the Melbourne dataset, where years of stable periodicity are available, supervised and foundation models substantially outperform Seasonal Naive, confirming that model complexity must match the available data budget.
Tree ensembles generalize well when the test distribution resembles training, as on medium- and low-volume sensors with stable daily profiles, but become unstable on high-volume sensors where crowd surges during later event days fall outside the axis-aligned partitions learned from earlier, calmer days.

Returning to the question that motivates this study, whether FMs deliver reliable gains over classical methods, the answer is regime-dependent.
FMs provide genuine cold-start value (Chronos-2 achieves the lowest 2-hour MAE across all sensors in the zero-shot setting E1), but this advantage erodes as local data accumulates.
In the seasonal regime, the FM advantage is driven by context length rather than inherent architectural superiority: short-context inference ($L=168$) performs comparably to supervised baselines, whereas extending context to 16{,}000 hours yields MAE reductions of approximately 16 to 49\% in the reported Melbourne results.
For deployment, a tiered strategy is therefore appropriate: Seasonal Naive as a long-horizon reference in limited-history settings, a lightweight tree model for short-horizon forecasting once one to two days have accumulated, zero-shot FMs during the initial hours of a new deployment, and long-context FMs for planning horizons in data-rich seasonal regimes.
The cost measurements reinforce this tiering: Seasonal Naive is not merely accurate at long horizons under limited history but also essentially free, LightGBM gives the best accuracy per unit of cost in the event regime at roughly a seventh of CatBoost's training time, and N-HiTS dominates the seasonal regime on both counts.
PatchTST, by contrast, is among the most expensive models to train in either regime yet rarely the most accurate, and is therefore hard to justify for deployments of this kind.

\section{Conclusions}
\label{sec:conclusion}

We benchmarked seven univariate forecasting methods across an event-driven, limited-history setting (SAIL2025) and a multi-year seasonal setting (Melbourne) to assess whether time-series foundation models deliver reliable gains over classical methods for pedestrian-count forecasting.
Model performance depends on the regime, and no single model family dominates under all conditions.
Under limited event history, Seasonal Naive and gradient-boosted trees remain competitive, while FMs provide cold-start value that diminishes as local data accumulates.
In the seasonal regime, long-context FMs achieve the best accuracy at weekly and monthly horizons, with MAE reductions of approximately 16 to 49\% in the reported Melbourne results when the TimesFM context is extended from 1 week to 16{,}000 hours.
Model selection should therefore be guided by available history and distributional shift rather than by a default preference for the most complex architecture.

Because our evaluation is univariate, all models' inability to capture schedule-driven crowd surges highlights the need for exogenous event features in operational forecasting.
Future work should incorporate such covariates, extend the cost profiling to context-length scaling, and extend the benchmark to calibrated uncertainty estimation for risk-aware crowd management.

\section*{ACKNOWLEDGMENTS}
This research is part of the AIM-TT learning community project, funded by the European Union--NextGenerationEU. The AI-COMPASS project also supports this work, funded by the Dutch Research Council (NWO) under KICH1.VE04.22.007.

	\bibliographystyle{IEEEtran}
	\bibliography{root} 
	
\end{document}